\documentclass[lettersize,journal]{IEEEtran}
\usepackage{amsmath,amsfonts}
\usepackage{algorithmic}
\usepackage{algorithm}
\usepackage{array}
\usepackage[caption=false,font=normalsize,labelfont=sf,textfont=sf]{subfig}
\usepackage{textcomp}
\usepackage{stfloats}
\usepackage{url}
\usepackage{verbatim}
\usepackage{graphicx}
\usepackage{cite}
\usepackage{stfloats}
\usepackage{enumitem}
\usepackage{multirow}
\usepackage{float}
\usepackage{caption}
\usepackage{subcaption}
\ifCLASSOPTIONcaptionsoff
 \usepackage[nomarkers]{endfloat}
\let\MYoriglatexcaption\caption
\renewcommand{\caption}[2][\relax]{\MYoriglatexcaption[#2]{#2}}
\fi
\usepackage{url}
\usepackage{hyperref}
\hypersetup{
    colorlinks=true,
    linkcolor=blue,
    citecolor=blue,      
    urlcolor=blue,
}

\begin{document}

\title{Noise-Augmented Boruta: The Neural Network Perturbation Infusion with Boruta Feature Selection}

\author{Hassan Gharoun\textsuperscript{1}, Navid Yazdanjoe\textsuperscript{2}, Mohammad Sadegh Khorshidi\textsuperscript{3}, and Amir H. Gandomi\textsuperscript{4,5,6}%
\thanks{\textsuperscript{1}Faculty of Engineering \& IT, University of Technology Sydney, email: Hassan.gharoun@student.uts.edu.au.}
\thanks{\textsuperscript{2}Faculty of Engineering \& IT, University of Technology Sydney, email: Navid.Yazdanjue@gmail.com}
\thanks{\textsuperscript{3}Faculty of Engineering \& IT, University of Technology Sydney, email: Mohammadsadegh.khorshidialikordi@student.uts.edu.au}
\thanks{\textsuperscript{4}Faculty of Engineering \& IT, University of Technology Sydney, email: Gandomi@uts.edu.au.}
\thanks{\textsuperscript{5}University Research and Innovation Center (EKIK), Óbuda University.}%
\thanks{\textsuperscript{6}Corresponding author}%
}



\maketitle

\begin{abstract}
With the surge in data generation, both vertically (i.e., volume of data) and horizontally (i.e., dimensionality) the burden of the curse of dimensionality has become increasingly palpable. Feature selection, a key facet of dimensionality reduction techniques, has advanced considerably to address this challenge. One such advancement is the Boruta feature selection algorithm, which successfully discerns meaningful features by contrasting them to their permutated counterparts known as shadow features. However, the significance of a feature is shaped more by the data's overall traits than by its intrinsic value, a sentiment echoed in the conventional Boruta algorithm where shadow features closely mimic the characteristics of the original ones. Building on this premise, this paper introduces an innovative approach to the Boruta feature selection algorithm by incorporating noise into the shadow variables. Drawing parallels from the perturbation analysis framework of artificial neural networks, this evolved version of the Boruta method is presented. Rigorous testing on four publicly available benchmark datasets revealed that this proposed technique outperforms the classic Boruta algorithm, underscoring its potential for enhanced, accurate feature selection. 
\end{abstract}


\section{Introduction}
\IEEEPARstart{W}{ith} the emergence of data centers and the advent of big data technologies in recent years, there has been a marked influence on the processes of data generation and storage, where these advancements have acted as powerful enablers for high-throughput systems, substantially augmenting the capacity to generate data both in terms of the number of data points (sample size) and the range of attributes or features collected for each data point (dimensionality) \cite{tang2014feature}. The explosive surge in the volume of gathered data has heralded unprecedented opportunities for data-driven insights. Yet, high-dimensionality simultaneously poses distinct challenges that obstruct the success of machine learning algorithms. This dichotomy is particularly emphasized in the so-called \textit{"curse of dimensionality"}, a term coined by Richard Bellman\cite{bellman1959adaptive}, which encapsulates the challenges faced in handling high-dimensional data spaces. The curse of dimensionality is effectively addressed by employing a collection of techniques collectively referred to as dimensionality reduction. Dimensionality reduction can be categorized into two primary branches: 
\begin{enumerate}[label=\alph*.]
    \item Feature extraction: the process of creating a smaller collection of new features from the original dataset while still preserving the majority of the vital information.
    \item Feature selection: the process of identifying and choosing the most relevant features from the original dataset based on their contribution to the predetermined relevance criterion. 
\end{enumerate}

Feature selection, similar to machine learning models, is classified into supervised, unsupervised, and semi-supervised types, depending on the availability of well-labeled datasets. Furthermore, supervised feature selection is divided into three main subcategories, namely (interested readers in delving deeper into feature extraction and feature selection, and their various types, are encouraged to refer to \cite{zebari2020comprehensive}): 
\begin{enumerate}[label=\roman*.]
    \item Filter methods: rank features based on statistical measures and select the top-ranked features.
    \item Wrapper methods: evaluates subsets of features which best contribute to the accuracy of the model
    \item Hybrid methods: leverages the strengths of both filter and wrapper methods by first implementing a filter method to simplify the feature space and generate potential subsets, and then using a wrapper method to identify the most optimal subset \cite{tang2014feature}.
    \item Embedded methods: utilize specific machine learning models that use feature weighting functionality embedded in the model to select the most optimal subset during the model's training \cite{jovic2015review}.
\end{enumerate}
Random Forest is a widely used algorithm for embedded feature selection. The Random Forest (RF) algorithm is a type of ensemble classifier that uses a concurrent set of decision trees, termed component predictors. RF applies a bootstrapping technique that randomly creates $n$ training subsets from the main dataset, and this process is performed $m$ times, leading to the construction of $m$ independent decision trees. Each tree is built using a random subset of features. The ultimate decision is made based on the majority vote of the component predictors \cite{habibpour2021uncertainty}. RF leverage permutation importance to calculate feature importance. Each tree in the forest contributes to the classification of instances it wasn't used to train. After the initial classification, the feature values are shuffled randomly and the classification process is repeated. The importance of a feature is determined by comparing the correct classifications before and after the permutation of feature values. If shuffling a feature's values leads to a large decrease in accuracy, then that feature is considered important. The final feature importance is obtained by averaging the importance of all individual trees. The utilization of Z-score is another viable approach, wherein the average value is divided by its standard deviation to derive an importance measure \cite{kursa2010boruta}. Algorithm \ref{Alg: RF feature importance} outlines the process of calculating feature importance in a RF model.

\begin{algorithm}
\caption{Calculate Feature Importance by RF}
\label{Alg: RF feature importance}
\begin{algorithmic}
\REQUIRE Random Forest, Instances, Features
\ENSURE Feature Importance
\STATE $V_{\text{orig}} \leftarrow 0$
\STATE $V_{\text{perm}} \leftarrow 0$
\FOR{each tree $t$ in Random Forest}
    \IF{tree $t$ did not use the current instance for training}
        \STATE Classify all instance
        \STATE $V_{\text{orig}}^{(t)} \leftarrow$ number of correct votes
        \STATE Permute feature values
        \STATE Classify all instance again
        \STATE $V_{\text{perm}}^{(t)} \leftarrow$ number of correct votes
        \STATE $I^{(t)} \leftarrow \frac{V_{\text{orig}}^{(t)} - V_{\text{perm}}^{(t)}}{\text{number of instances}}$
    \ENDIF
\ENDFOR
\STATE $I \leftarrow \frac{1}{\text{number of trees}} \sum_{t=1}^{\text{number of trees}} I^{(t)}$
\RETURN $I$
\end{algorithmic}
\end{algorithm}

\cite{kursa2010boruta} argued that the trustworthiness of the evaluation of feature significance is grounded in the presumption that the separate trees cultivated within the random forest are unrelated while numerous analyses have occasionally demonstrated that this presupposition might not hold true for certain datasets. Furthermore, they contended that distinguishing genuinely important features becomes difficult when dealing with a large number of variables, as some may seem important due to random data correlations. Accordingly, the importance score by itself is inadequate to pinpoint significant associations between features and the target\cite{kursa2010boruta}. They address this issue by proposing Boruta algorithm. 

In Random Forest, the importance of features is calculated in comparison to each other. However, in Boruta, the main idea is to evaluate the importance of features in competition with a set of random features called shadow features. In this process, every feature in the dataset is duplicated and their values are shuffled randomly. The Random Forest algorithm is applied repeatedly, randomizing the shadow features each time and calculating feature importance for all attributes (original features and shadow features). If the importance of a given feature consistently exceeds the highest importance among all the shadow features, it is classified as important. The measure of consistency is established through a statistical test, grounded on the binomial distribution, which quantifies how frequently the feature's importance overtakes the Maximum Importance of the Random Attributes (MIRA). If this count (called 'hits') significantly outnumbers or undershoots the expected count, the feature is deemed 'important' or 'unimportant' respectively. This process iterates until either all features are conclusively categorized, or a predetermined iteration limit is reached. Algorithm \ref{Alg: Boruta} succinctly illustrates the steps of the Boruta algorithm \cite{kursa2010boruta}.

\begin{algorithm}
\caption{Boruta Algorithm for Feature Selection}
\label{Alg: Boruta}
\begin{algorithmic}
\STATE Let $\mathcal{F}$ be the set of all features
\STATE Let $\mathcal{H}$ be the empty list to store the importance history
\STATE Let $maxIter$ be the maximum number of iterations
\FOR {$iter=1$ to $maxIter$}
    \STATE Create $\mathcal{F'} = \mathcal{F} \cup \{\text{shuffled~copies~of~all~} f \in \mathcal{F}\}$ (the shadow features)
    \STATE Train a $RF$ classifier on the dataset using $\mathcal{F'}$
    \STATE Compute $I = RF.\text{importance}$, the importance score for all features in $\mathcal{F'}$
    \STATE Compute $maxShadow = \max(I_{f'})$ where $f' \in \mathcal{F'}$ are the shadow features
    \FOR {each $f \in \mathcal{F}$}
        \STATE Add $I_f$ to $\mathcal{H}_f$, the importance history for feature $f$
        \IF {$\bar{I}_f > maxShadow$}
            \STATE Mark $f$ as important
        \ELSIF {$\bar{I}_f < maxShadow$ for some threshold number of times in $\mathcal{H}_f$}
            \STATE Mark $f$ as unimportant
        \ENDIF
    \ENDFOR
\ENDFOR
\RETURN The set of features marked as important
\end{algorithmic}
\end{algorithm}

Since the introduction of Boruta, this algorithm has been extensively and successfully utilized in across diverse research domains, including medicine\cite{tang2020cart, maurya2023prognostic, Debbi2023, SantosFebles2022}, cybersecurity\cite{subbiah2022intrusion}, engineering\cite{RashidiNasab2023, Maliuk2022}, and environmental\cite{heidari2022wavelet, ahmadpour2021gully, jamei2023high, Subbiah20233829} studies. Even the Boruta algorithm has been successfully employed to reduce the dimensionality of features extracted from images by deep networks\cite{Borugadda2023813}.
While the Boruta algorithm has indeed been successful in feature selection, contributing to improved predictive performance as highlighted in the literature, it's crucial to note that in Boruta, features are merely permuted. This permutation does not alter the inherent attributes of a feature. A similar phenomenon occurs in the Random Forest algorithm when calculating feature importance through permutation. However, The relevance of the feature is determined by the data's characteristics, not its value \cite{venkatesh2019review}. Therefore, in this study, we introduce a new variant of the Boruta algorithm. In the traditional Boruta algorithm, shadow features are constructed merely by random shuffling, which does not alter a feature's properties. To address this, in our study, the shadow features are not only shuffled but also perturbed with normal noise. Additionally, instead of employing RF for calculating feature importance, we have utilized the perturbation analysis approach within neural networks. \\
The remainder of this paper is structured as follows: Section \ref{section: proposed method} offers a comprehensive discussion of the proposed algorithm. Section \ref{section:Experimental Setup} details the dataset used and outlines the experimental design. The findings from the experiments are presented and analyzed in Section \ref{section:Discussion and Results}. Lastly, Section \ref{section:Conclusion} provides concluding remarks and suggests avenues for future research.

\section{Proposed Method}
\label{section: proposed method}
\subsection{Noise-augmented shadow features}
The value of a feature in a dataset is often viewed as less important than the overall characteristics of the data, in terms of providing insight into the predictive modeling process \cite{venkatesh2019review}. This perspective holds true for the Boruta algorithm, where the shadow features, bearing the same characteristics as the original ones, are utilized. Yet, it should be noted that even though the permutation of these shadow features disrupts the original relationship with the target variable, the essence of the Boruta algorithm—that each original feature competes with random features mirroring their own characteristics—remains intact.\\
This study aims to further the current understanding of the role and potential of shadow features by questioning the foundational assumption of their inherent similarity to the original features. To this end, the concept of 'noise-augmented shadow features' is proposed, where the characteristics of the shadow features are deliberately modified. This new approach allows for an exploration into whether diversifying the characteristics of shadow features can lead to improved feature selection performance. The theoretical rationale behind this new approach is to provide a broader spectrum of random features for the original ones to compete against, thereby enriching the competition space and potentially enhancing the robustness of the feature selection process. This investigation is driven by the belief that the performance of a feature selection algorithm may be influenced not only by the relevancy of the features but also by the diversity and characteristics of the shadow features. \\
Algorithm \ref{Alg: NoiseShadow} clearly outlines the steps involved in generation of noise-augmented shadow features. In this approach, each original feature undergoes a process of augmentation with white noise – a random value possessing zero mean and standard deviation equal to that computed from the original feature. This noise generation step mimics the statistical characteristics of the original feature while simultaneously disrupting its inherent relationship with the target variable. Subsequently, a random permutation is applied, which further ensures that any patterns or dependencies present in the original feature set do not unduly influence the shadow features.

\begin{algorithm}
\caption{Generation of Noise-Augmented Shadow Features}
\label{Alg: NoiseShadow}
\begin{algorithmic}[1]
\STATE Let $F$ be the set of all features
\REQUIRE $F$ 
\FOR{each feature $f$ in $F$}
    \STATE $\delta \leftarrow$ compute standard deviation of feature $f$
    \STATE $Noise \leftarrow$ generate white noise with 0 mean and $\delta$
    \STATE $Shadow_{i} \leftarrow$ augment feature $f$ with $Noise$ then permute randomly
\ENDFOR
\RETURN $F_{NS}$: Noise-augmented shadow features.
\end{algorithmic}
\end{algorithm}

\subsection{Perturbation-based assessment of feature importance}
\begin{figure}[!htbp]
  \centering
  \includegraphics[width=\columnwidth]{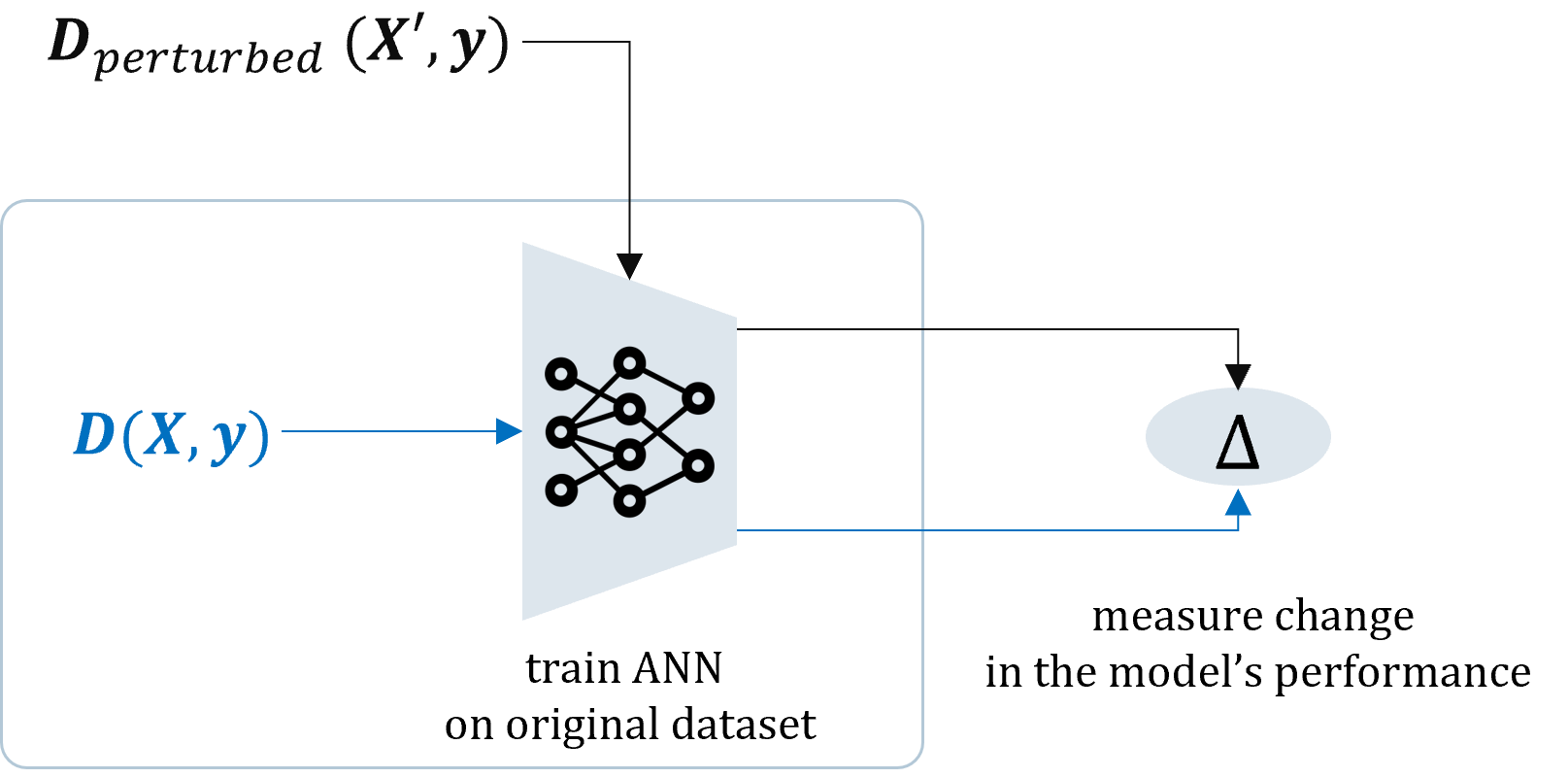}
  \caption{Perturbation analysis scheme.}
  \label{fig:Perturbation analysis}
\end{figure}

The concept of perturbation analysis offers a solution to quantify the influence of each variable within the framework of neural network models. In the procedure, disturbances are intentionally introduced to the neural network's inputs. To maintain control over the experiment, only one input variable is altered during each iteration, keeping the remainder unchanged. The variable that, when disturbed, yields the most significant impact on the dependent variable is then recognized as the variable of greatest importance\cite{gharoun2019integrated}. Figure \ref{fig:Perturbation analysis} shows the general scheme of perturbation analysis.

In light of the above, this study introduces a novel variant of the Boruta feature selection method, inspired by the perturbation analysis paradigm employed in Artificial Neural Networks (ANNs). This approach incorporates the use of noise-augmented shadow features and utilizes a Shallow ANN as the underlying base model.\\
Let's consider a dataset, $D = {(x_1, y_1), (x_2, y_2), ..., (x_N, y_N)}$, where $x_i$ represents the $i^{th}$ observation vector in a $d$-dimensional feature space, and $y_i$ corresponds to the label of the $i^{th}$ observation. The first stage involves the creation of training and testing datasets, denoted by $D_{train}$ and $D_{test}$, respectively. \\
In this proposed variant, $D_{train}$ is solely used for feature selection, while $D_{test}$ is reserved exclusively to evaluate the performance of the selected features. Thus, the feature selection process does not have any access to or influence from the test dataset, thereby ensuring an unbiased assessment of the feature selection process.\\
Algorithm \ref{Alg: Noise-augmented Boruta} offers a step-by-step delineation of the proposed method.

\begin{algorithm}
\caption{Noise-augmented Boruta}
\label{Alg: Noise-augmented Boruta}
\begin{algorithmic}[1]
\STATE Let $D_{train}$ be the train set with feature set F
\STATE Let $D_{NS}$ be the set of noise-augmented features set
\STATE Let $\mathcal{H}$ be the empty list to store the hit history 
\STATE Let $maxIter$ be the maximum number of iterations
\REQUIRE Shallow ANN, $D_{train}$, $D_{NS}$, $maxIter$
\FOR{$iter=1$ to $maxIter$}
    \STATE Create $\mathcal{D'} = D_{train} \cup D_{NS}$
    \STATE Normalize $\mathcal{D'}$
    \STATE $Model \leftarrow$ Shallow ANN
    \STATE Train the model on the dataset $\mathcal{D'}$
    \STATE Compute $I \leftarrow$ the training f1 score of model on $\mathcal{D'}$ 
    \FOR{each feature $f$ in $\mathcal{D'}$}
        \STATE Perturb \& shuffle feature $f$ by adding ($1+n\sigma$) while keeping other features unchanged
        \STATE Compute ${I}'_f \leftarrow$ the f1 score of trained model on $\mathcal{D'}$ with perturbed feature $f$
        \STATE ${I}''_f \leftarrow \max(I - I', 0)$
    \ENDFOR
    \STATE Normalizing every ${I}''_f$ by dividing to sum of all $I''$
    \STATE $I_{Max Shadow} \leftarrow \max(I'')$ among $D_{NS}$
    \FOR{every feature $f$ in $D_{train}$ (original feature)}
        \IF{${I}''_f > I_{Max shadow}$}
            \STATE Add a hit to $\mathcal{H}_f$, the hit history for feature $f$
        \ENDIF
    \ENDFOR
\ENDFOR
\RETURN The set of features with at least one hit 
\end{algorithmic}
\end{algorithm}

In the proposed method, given $D_{train}$ a new train set $\mathcal{D'}$ is constructed by a combination of original features and their noise-augmented counterparts (shadow features). This set is then normalized to prepare it for the learning algorithm (ANN shallow learner). The F1 score of the trained model by $\mathcal{D'}$ is then used as the baseline performance metric. Next, each feature in the $\mathcal{D'}$ is perturbed individually by adding a noise factor and shuffling while keeping the other features unchanged. The perturbed F1 score of the model is computed and the difference between the baseline and perturbed scores is noted. The difference in scores effectively quantifies the influence of perturbing each feature, and these differences are then normalized. This step allows us to measure the influence of each feature on the model's performance. Afterward, a competition takes place between the original features and their noise-augmented shadow counterparts. The most influential shadow feature (i.e., the one with the highest normalized difference in F1 score after perturbation) sets a threshold. If an original feature's impact on the F1 score exceeds this threshold, it is considered important and one \textit{'hit'} is recorded. This process is repeated for a pre-specified number of iterations. At the end of these iterations, the features that have accumulated at least one hit are selected.

\section{Experimental Setup}\label{section:Experimental Setup}
\subsection{Data sets}
To evaluate the proposed method's performance, it was applied to four publicly available datasets. These datasets, each unique in their characteristics, offer a diverse range of scenarios to thoroughly test the robustness and efficacy of the proposed method. Brief descriptions of each dataset are presented below:
\begin{itemize}
    \item Smartphone-based recognition of human activities and postural transitions (SB-RHAPT)\cite{SPRHAPT}: comprises of data collected from smartphone sensors recording basic activities and postural transitions. Each record is associated with an activity encoded as six different classes.
    \item APS Failure at Scania Trucks (APSF)\cite{apsf}: consists of data collected from the Air Pressure system (APS) of Scania Trucks. This dataset is imbalanced as most records represent normal operation while a small fraction corresponds to the APS failure. Missing values within this dataset are replaced with the mean value of the respective feature. 
    \item Epileptic seizure recognition (ESR)\cite{esr}: constitutes recorded EEG values over a certain period, aiming to distinguish between the presence and absence of epileptic seizure activity. The original target variable involves 5 different categories of brain activity patterns, four of them correspond to non-seizure activity and one category indicates seizure activity. In this study, the target variable is converted into a binary classification task to discern between seizure and non-seizure activities. This simplification leads to an imbalance in the dataset.
    \item Parkinson's disease classification (PDC)\cite{pdc}: incorporates instances representing various biomedical voice measurements from individuals, some of whom are afflicted with Parkinson's Disease. The dataset, designed for binary classification, categorizes instances into two classes of Parkinson's Disease and Healthy.
\end{itemize}
Summarized information about the utilized datasets, including the number of instances, features, and classes, can be found in Table \ref{tab:dataset}.

\begin{table}[]
\caption{Summary of datasets}
\label{tab:dataset}
\resizebox{\columnwidth}{!}{%
\begin{tabular}{lccc}
\hline
Dataset                                            & Instances & Features & Classes \\ \hline
Recognition of Human Activities & 10299     & 561      & 6       \\
Failure at Scania Trucks                           & 76000     & 171      & 2       \\
Epileptic Seizure Recognition                      & 11500     & 179      & 2       \\
Parkinson's Disease (PD) classification            & 756       & 755      & 2       \\ \hline
\end{tabular}%
}
\end{table}

\subsection{Experiment configurations}
In this study, the performance of the proposed method has been compared with the original Boruta algorithm. For feature selection using the Boruta algorithm, Random Forest is utilized. Two principal parameters used in this study to tune the Random Forest are the number of estimators and the maximum depth. To obtain the optimum value for these two parameters, a Random Forest was initially trained on each dataset with all features, and the best value was determined via a greedy search. Table \ref{tab:RF configuration} presents the optimal parameter values for Boruta's estimators across each dataset. The Random Forest model obtained at this stage is employed in the Boruta algorithm for feature selection. \\
In configuring the method proposed in this study, more parameters need to be decided upon. The first set of these parameters pertains to the learner model based on the artificial neural network. Given that in the proposed methodology, the learner is solely used for feature selection, and features are chosen based on the impact their perturbation has on reducing model accuracy, thus fine-tuning the learner at this stage is not critical. What is required here is to select a network architecture that can generate a minimum accuracy above 50 percent. Therefore, through trial and error, simple models capable of achieving an accuracy above 50 percent with all features are utilized. The chosen architecture can vary for each dataset based on the dataset's inherent characteristics. Table \ref{tab:Shallow learner configuration} depicts the architecture employed for each dataset. For all models, the epoch is set to 100. An observation that can be made from this table is that most models are very lightweight, which contributes to reducing the problem's complexity.\\
The subsequent parameter, denoted as 'n', serves as the standard deviation multiplier during the perturbation of parameters to assess the degree of accuracy reduction in the model. In this study, an 'n' value of 50 has been adopted.

\begin{table}[]
\caption{Boruta estimator configuration}
\label{tab:RF configuration}
\resizebox{\columnwidth}{!}{%
\begin{tabular}{ll}
\hline
Dataset                                                                                                                       & Random Forest                                                                          \\ \hline
Smartphone-Based Recognition of Human   Activities                                                                            & (200, None)                                                                                       \\
Failure at Scania Trucks                                                                                                      & (200, None)                                                                                      \\
Epileptic Seizure Recognition                                                                                                 & (200, None)                                                                                   \\
Parkinson's Disease (PD) classification                                                                                       & (200, 10)                                                                                       \\ \hline
\end{tabular}%
}
\par
\smallskip
\noindent
\footnotesize{Note:The sequence of numbers ($i_1$,$i_2$) presents the number of estimators and max depth repsectively.}
\end{table}

\begin{table}[]
\caption{Shallow learner configuration}
\label{tab:Shallow learner configuration}
\resizebox{\columnwidth}{!}{%
\begin{tabular}{ll}
\hline
Dataset                                                                                                                       & Shallow learner                                                                           \\ \hline
Smartphone-Based Recognition of Human   Activities                                                                            & (5)                                                                                       \\
Failure at Scania Trucks                                                                                                      & (16)                                                                                      \\
Epileptic Seizure Recognition                                                                                                 & (5,8,5)                                                                                   \\
Parkinson's Disease (PD) classification                                                                                       & (5)                                                                                       \\ \hline
\end{tabular}%
}
\par
\smallskip
\noindent
\footnotesize{Note: Each number signifies the size of neurons in a layer. In the cases where a sequence of numbers is presented, such as ($i_1$,$i_2$,$i_3$), these correspond to multiple hidden layers within the network.}
\end{table}

\section{Discussion and Results}\label{section:Discussion and Results}
This section presents the numerical results and discussion. As mentioned in the previous sections, the evaluation of the model is based on the F1 score, which provides a more robust measure in scenarios involving imbalanced datasets. Each of the datasets has been randomly divided into training and testing sets at a ratio of 70\% to 30\%, with stratified sampling from the target variable. The proposed method and the original Boruta algorithm are each run on the training dataset, with a maximum iteration limit of 100 times. The selected features are then used for the final evaluation of the selected features. To this end, the training set used in feature selection is filtered down to the selected features and retrained, and then evaluated on the test set. It is important to note that the test set is never exposed to the model at any stage of the training. \\
It should be noted that in the proposed method, similarly ANN -or to be more specific multi-layer perceptron (MLP) - is employed for evaluation of the selected features on the test set. In this stage, against the feature selection stage, it is necessary to fine-tune the neural network for evaluating the derived feature set. Table \ref{tab:mlp configuration} shows the architecture of the tuned network for each dataset. Here, the epoch is set to 1000.\\ The evaluation results are shown in Table \ref{tab:numerical_result} for the proposed method after fine-tuning MLP. \\
From an initial observation of the results, it can be inferred that the Noise-augmented Boruta consistently outperforms the standard Boruta in terms of F1 score across all datasets. Notably, this improvement is achieved with a significantly reduced number of selected features in most cases, indicating a more efficient feature selection by the Noise-augmented Boruta.

To gain a more robust understanding of performance variability - considering the inherent randomness in MLP (e.g., weight initialization) and RF (e.g., random subsamples) - each model was run 100 times. In each run, the entire dataset, filtered to include only the selected features, was randomly split into training and testing subsets. This procedure can be likened to K-fold cross-validation but with a higher number of repetitions. It allows the model to experience various potential distributions within the dataset, thus bolstering its robustness against unseen distributions. Furthermore, it captures the effects of variability originating from inherent randomness within the algorithm's performance. Table \ref{tab:100 run result} summarizes the performance results from the 100 runs, while Figure \ref{fig:ablation_boxplot} illustrates the distribution of the evaluation metric (F1 score) for both the proposed method and Boruta across the four datasets.

\begin{table}[]
\caption{Comparison result of the proposed method with Boruta}
\label{tab:numerical_result}
\resizebox{\columnwidth}{!}{%
\begin{tabular}{lccccc}
\hline
\multirow{2}{*}{Dataset} & \multirow{2}{*}{All Features} & \multicolumn{2}{c}{Boruta} & \multicolumn{2}{c}{Noise-augmented Boruta} \\ \cline{3-6} 
                         &                               & Sel. Feat.    & F1score    & Sel. Feat.            & F1score            \\ \hline
SB-RHAPT                 & 561                           & 479           & 92.407\%   & 104                   & 94.283\%           \\
APSF                     & 171                           & 55            & 86.820\%   & 22                    & 89.008\%           \\
ESR                      & 178                           & 178           & 95.067\%   & 138                   & 96.418\%           \\
PDC                      & 755                           & 78            & 77.802\%   & 29                    & 81.908\%           \\ \hline
\end{tabular}%
}
\end{table}

\begin{table}[]
\caption{MLP optimal architect}
\label{tab:mlp configuration}
\resizebox{\columnwidth}{!}{%
\begin{tabular}{ll}
\hline
Dataset                                                                                                                       & MLP                                                                           \\ \hline
Smartphone-Based Recognition of Human   Activities                                                                            & (512,512,256)                                                                                      \\
Failure at Scania Trucks                                                                                                      & (64,256,64)                                                                                      \\
Epileptic Seizure Recognition                                                                                                 & (128, 512, 128)                                                                                  \\
Parkinson's Disease (PD) classification                                                                                       & (1024, 1024, 512)                                                                                      \\ \hline
\end{tabular}%
}
\par
\smallskip
\noindent
\end{table}

\begin{table}[]
\caption{Comparison result of the proposed method with Boruta - 100 times run}
\label{tab:100 run result}
\resizebox{\columnwidth}{!}{%
\begin{tabular}{lcccc}
\hline
\multicolumn{1}{c}{\multirow{2}{*}{Dataset}} & \multicolumn{2}{c}{Boruta}             & \multicolumn{2}{c}{Noise-augmented Boruta} \\ \cline{2-5} 
\multicolumn{1}{c}{}                         & Sel. Feat. & F1score (\%)            & Sel. Feat.    & F1score (\%)             \\ \hline
SB-RHAPT                                     & 479               & 98.0886±0.30684 & 104                  & 98.8012±0.2209   \\
APSF                                         & 55                & 87.2672±0.7972  & 22                   & 87.6904±0.9294   \\
ESR                                          & 178               & 95.0774±0.4491  & 138                  & 95.8550±0.4358   \\
PDC                                          & 78                & 80.2250±3.0855  & 29                   & 81.1630±2.9937   \\ \hline
\end{tabular}%
}
\end{table}

\begin{table}[]
\caption{Statistical comparison of proposed method with Boruta}
\label{tab:stat_comparison}
\resizebox{\columnwidth}{!}{%
\begin{tabular}{lcccc}
\hline
\multirow{2}{*}{} & \multicolumn{2}{c}{Shapiro–Wilk test} & \multirow{2}{*}{Two-sample t-test} & \multicolumn{1}{l}{\multirow{2}{*}{Mann-Whitney U}} \\ \cline{2-3}
                  & Boruta          & NB     &                                    & \multicolumn{1}{l}{}                                     \\ \cline{2-5} 
SB-RHAPT          & 0.4029990      & 0.32806221          & 0.000000000                        & -                                                        \\
APSF              & 0.7328069      & 0.5650118          & 0.000670263                        & -                                                        \\
ESR               & 0.5187594      & 0.0002063          & -                                  & 0.000000000                                              \\
PDC               & 0.1229854      & 0.8666141          & 0.030293408                        & -                                                        \\ \hline
\end{tabular}%
}
\end{table}

The comparative analysis displayed in Table \ref{tab:100 run result} convincingly establishes the remarkable superiority of the Noise-augmented Boruta method over the conventional Boruta approach. Examining each dataset, it becomes apparent that the proposed method is more effective in the elimination of redundant or non-essential features, consistently selecting a significantly smaller feature set. Fewer feature set leads to simpler, less complex models that offer more interpretable results and reduce computational demand. Most impressively, this winnowing process does not compromise model performance. In fact, the Noise-augmented Boruta method equals or surpasses the F1 score achieved by the traditional Boruta across all tested datasets. The improvement is clear, from an increase in the F1 score on the SB-RHAPT dataset from 98.0886\% to 98.8012\%, and a rise on the PDC dataset from 80.2250\% to 81.1630\%. Even in instances like the APSF and ESR datasets, where the F1 score sees only slight growth, the proposed method proves its robustness, maintaining competitive performance despite the substantial reduction in the number of features. It is worth mentioning that the Noise-augmented Boruta method yielded lower variance in the F1 scores compared to the traditional Boruta method.

Building on this comparative analysis, Table \ref{tab:stat_comparison} presents the results of a rigorous statistical analysis comparing the proposed method and the Boruta method. This is based on the outcomes of 100 runs for each method across four distinct datasets: SB-RHAPT, APSF, ESR, and PDC. 
The Shapiro-Wilk test was first applied to check for normality in the distribution of results. For three out of the four datasets—SB-RHAPT, APSF, and PDC—the p-values observed in the Shapiro-Wilk test for both methods exceeded the 0.05 threshold, indicating a reasonable assumption of normal distribution. Therefore, the two-sample t-test was employed for these datasets. However, for the ESR dataset, the proposed method's results deviated from a normal distribution, as evidenced by its p-value of 0.0002063. As a result, the Mann-Whitney U test was employed as an appropriate non-parametric alternative for this dataset.

Across all datasets, the p-values resulting from the comparative tests were significantly below the 0.05 level, reinforcing that the performance of the two methods is distinct and statistically significant. Furthermore, as the proposed method consistently yielded higher accuracies, it can be concluded that the proposed method outperforms the Boruta method in the considered datasets.

\begin{figure}[!htbp]
  \centering
  \includegraphics[width=\columnwidth]{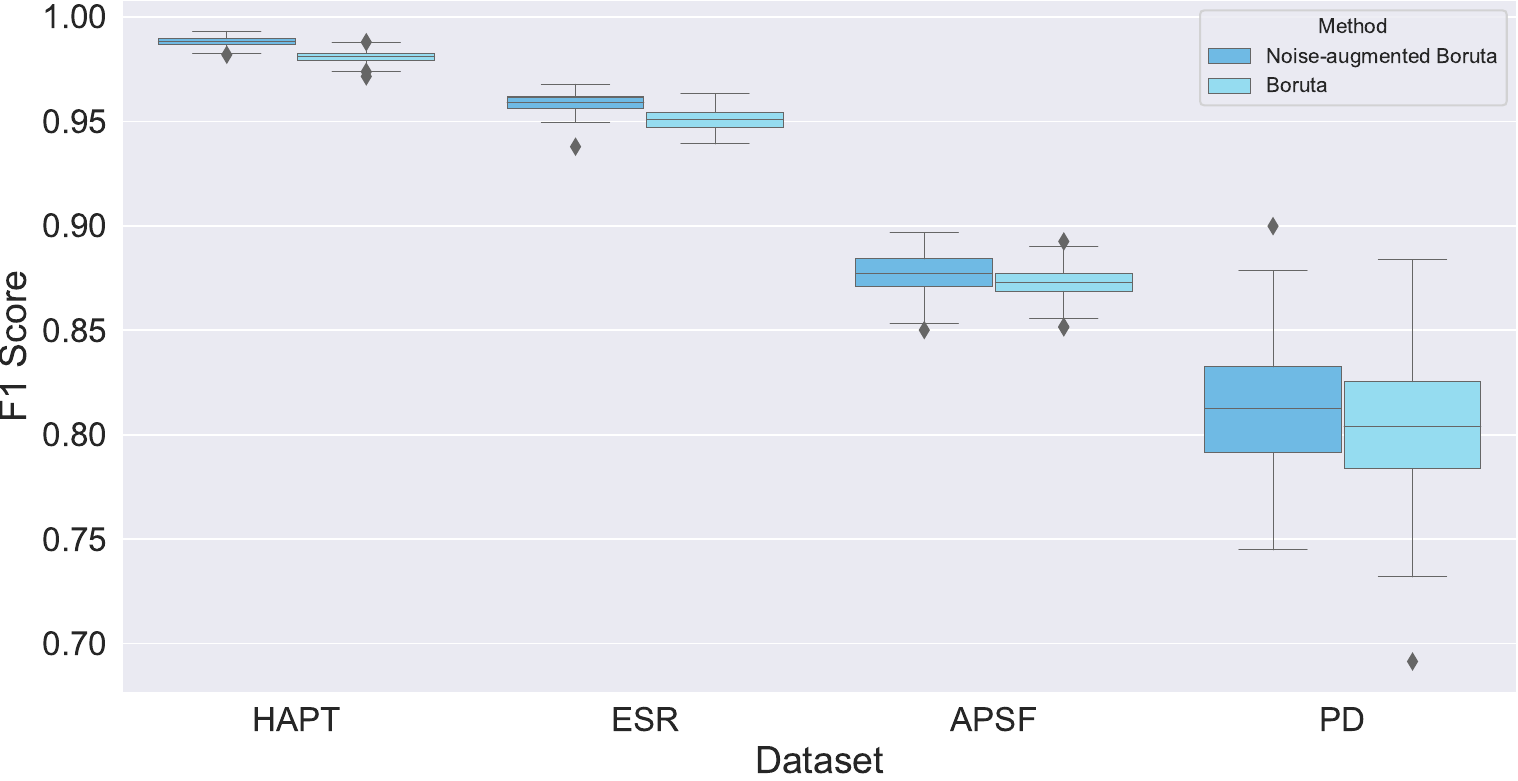}
  \caption{Box-plots of f1 score for the proposed method and Boruta.}
  \label{fig:Noise_augmented_boxplot}
\end{figure}

\begin{figure}[!t]
  \centering
  \captionsetup[subfloat]{font=tiny}
  \subfloat[][B; SB-RHAPT]{\includegraphics[width=.5\columnwidth]{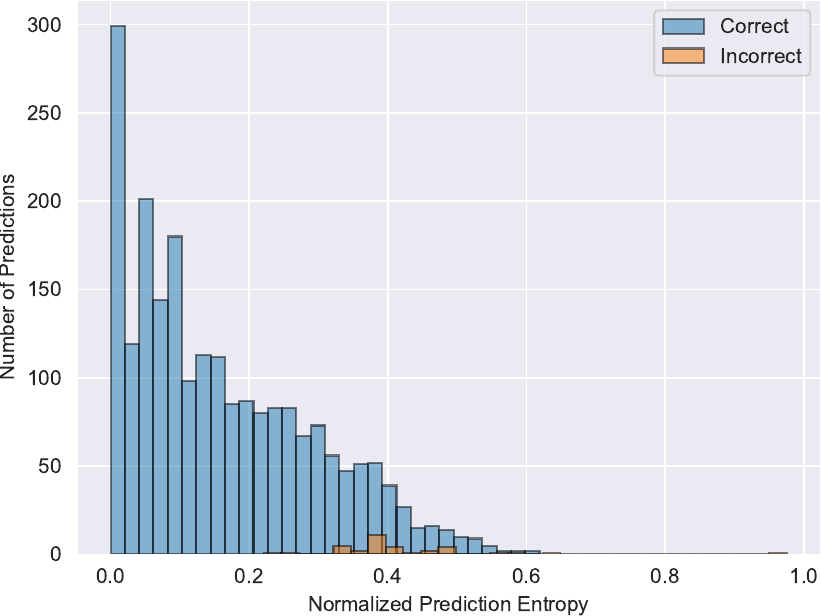}}\hfill
  \subfloat[][NB; SB-RHAPT]{\includegraphics[width=.5\columnwidth]{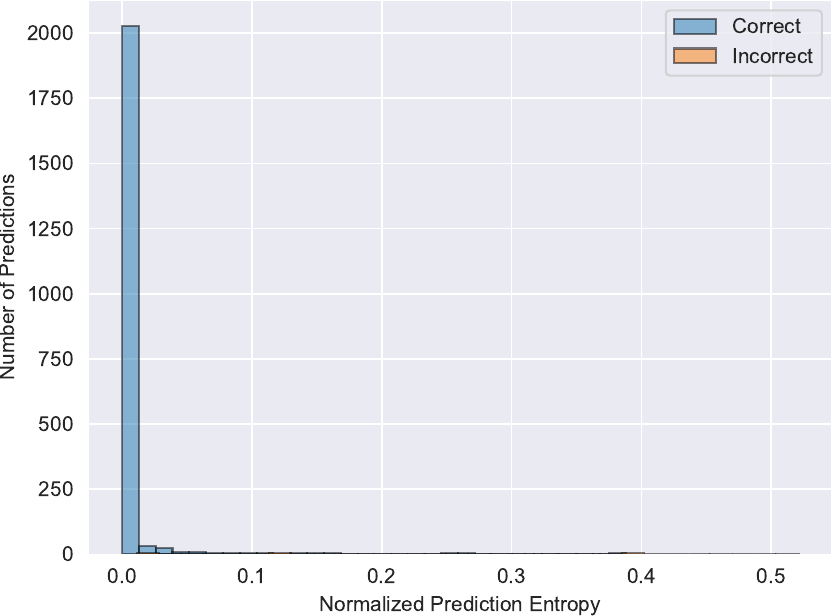}}\par
  \subfloat[][B; ESR]{\includegraphics[width=.5\columnwidth]{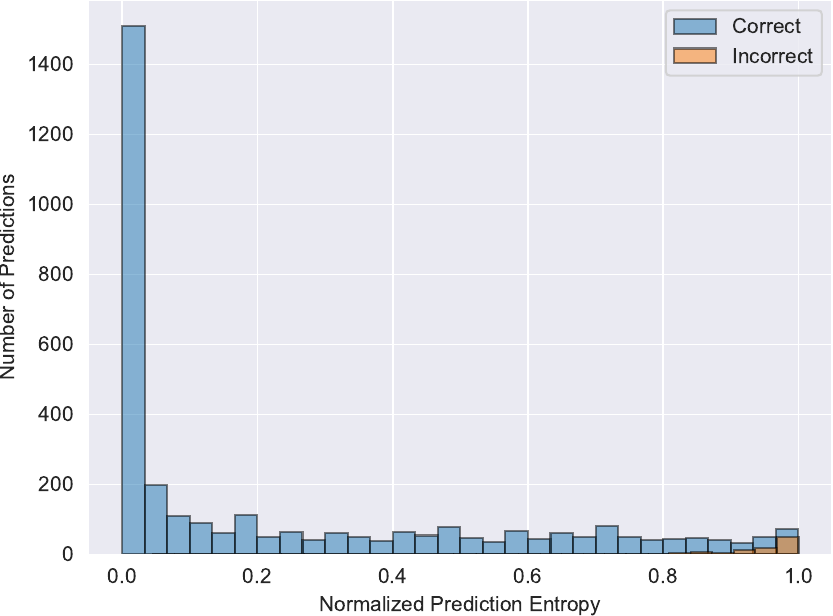}}\hfill
  \subfloat[][NB; ESR]{\includegraphics[width=.5\columnwidth]{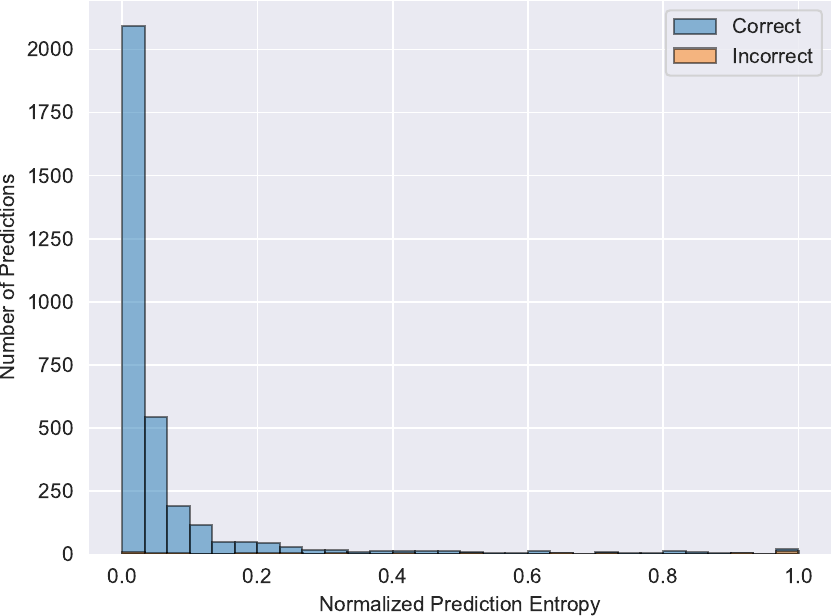}}\par
  \subfloat[][B; APSF]{\includegraphics[width=.5\columnwidth]{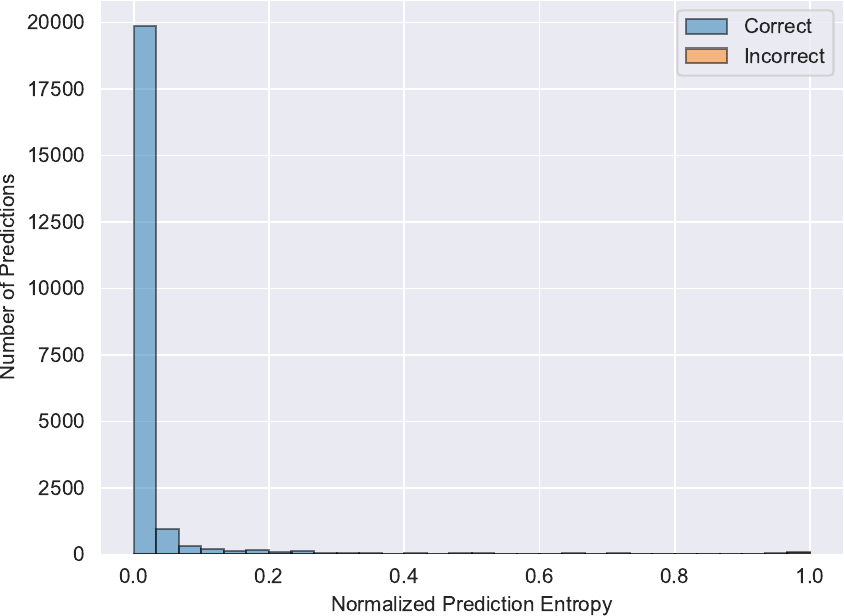}}\hfill
  \subfloat[][NB; APSF]{\includegraphics[width=.5\columnwidth]{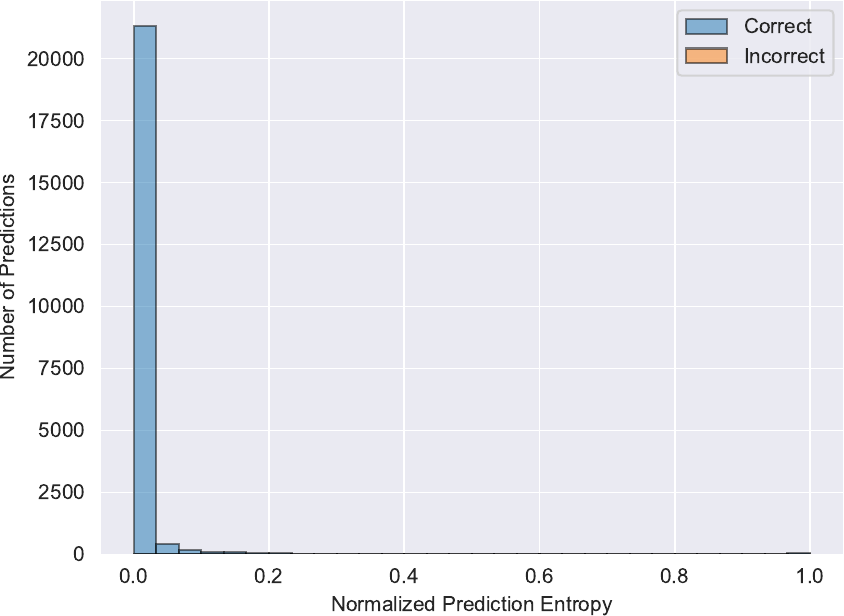}}\par
  \subfloat[][B; PDC]{\includegraphics[width=.5\columnwidth]{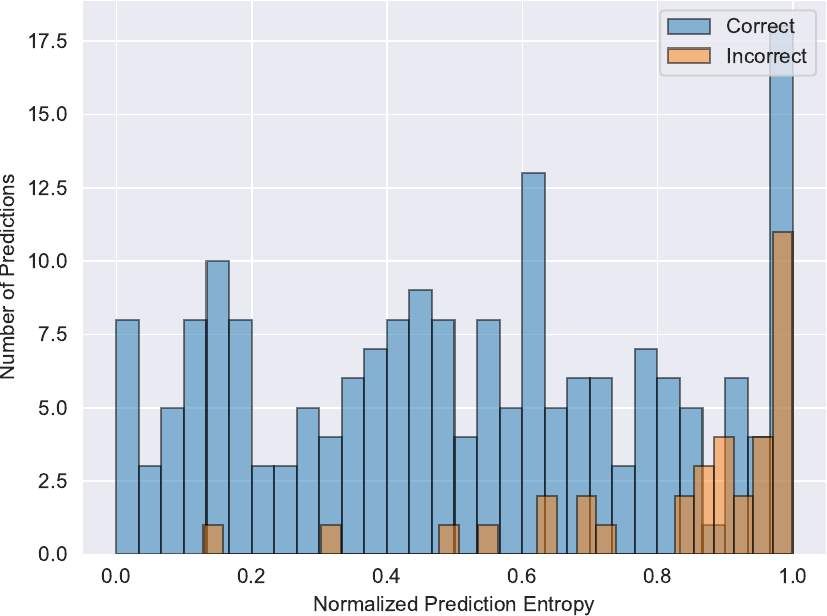}}\hfill
  \subfloat[][NB; PDC]{\includegraphics[width=.5\columnwidth]{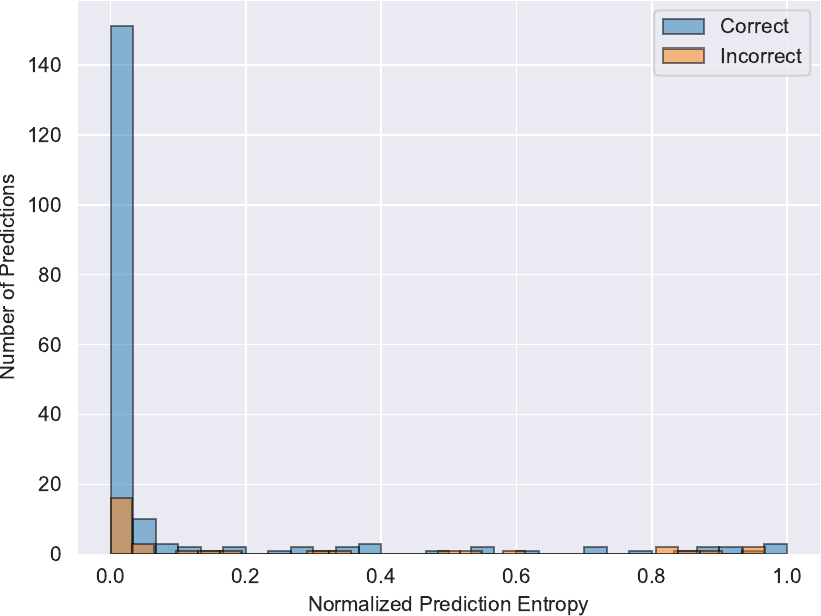}}\par
  \caption{Histogram graphs of the predictive entropy results. B and NB indicate Boruta and Noised-augmented Boruta respectively.}
  \label{fig:PE}
\end{figure}

For a deeper understanding of the comparison between the two models, Figure \ref{fig:PE} offers detailed insights into the models' confidence levels, as assessed by their prediction entropy. The prediction entropy for every instance $x_j$ in test set is calculated as \ref{eq:PE} \cite{habibpour2023uncertainty}:

\begin{equation}
    H(x_j) = - \sum_{i=1}^{n} p(c_i | x_j) \log_2 p(c_i | x_j)
\label{eq:PE}
\end{equation}
where:
$H(x_j)$ represents the entropy for the $j^{th}$ instance, $p(c_i | x_j)$ and denotes the probability that instance $x_j$ belongs to class $c_i$ ($i = 1, ..., n$).\\
The prediction entropy is normalized by dividing by the maximum possible entropy, making the results range between 0 (indicating complete certainty in prediction) and 1 (indicating complete uncertainty). A prediction with lower entropy means the model is more certain of its decision, while a higher entropy suggests the opposite. To elucidate the relationship between prediction confidence and accuracy, Figure \ref{fig:PE} separates and visualizes the distribution of entropies for both correctly and incorrectly predicted samples. This segregation provides a valuable perspective: if, for instance, incorrect predictions predominantly have high entropy, it indicates that the model is generally unsure when it errs. On the other hand, if incorrect predictions have low entropy, it suggests that the model is confidently making those mistakes. Figure \ref{fig:PE} provides evidence of the proposed method's superiority, showcasing its enhanced confidence across all four datasets compared to the Boruta algorithm.

\begin{table*}[t]
\caption{Ablation analysis results on $n$}
\label{tab:Ablation}
\resizebox{\textwidth}{!}{%
\begin{tabular}{lclcclclcl}
\hline
\multirow{2}{*}{Dataset} & \multirow{2}{*}{Number of all Features} & \multicolumn{2}{c}{n = 5}                        & \multicolumn{2}{c}{n =   20}                     & \multicolumn{2}{c}{n =   50}                     \\ \cline{3-8} 
       &              & Sel. Feat. & F1 score (\%) & Sel. Feat. & F1 score (\%) & Sel. Feat. & F1 score (\%) \\ \hline
SB-RHAPT &        561      & 119               & 99.0087±0.2428            & 117               & 98.8782±0.2288            & 104               & 98.8012±0.2209            \\
APSF     &    171          & 8                 & 79.8743±2.0264            & 23                & 86.7719±1.2292            & 22                & 87.6904±0.9294            \\
ESR      &        178      & 31                & 95.5689±0.4350            & 112               & 95.8388±0.4324            & 138               & 95.8550±0.4358            \\
PDC      &       755       & 36                & 81.7975±2.6048            & 25                & 79.3420±2.6942            & 29                & 81.1630±2.9937            \\ \hline
\end{tabular}%
}
\end{table*}

\subsection{Ablation study}
\begin{figure}[!htbp]
  \centering
  \includegraphics[width=\columnwidth]{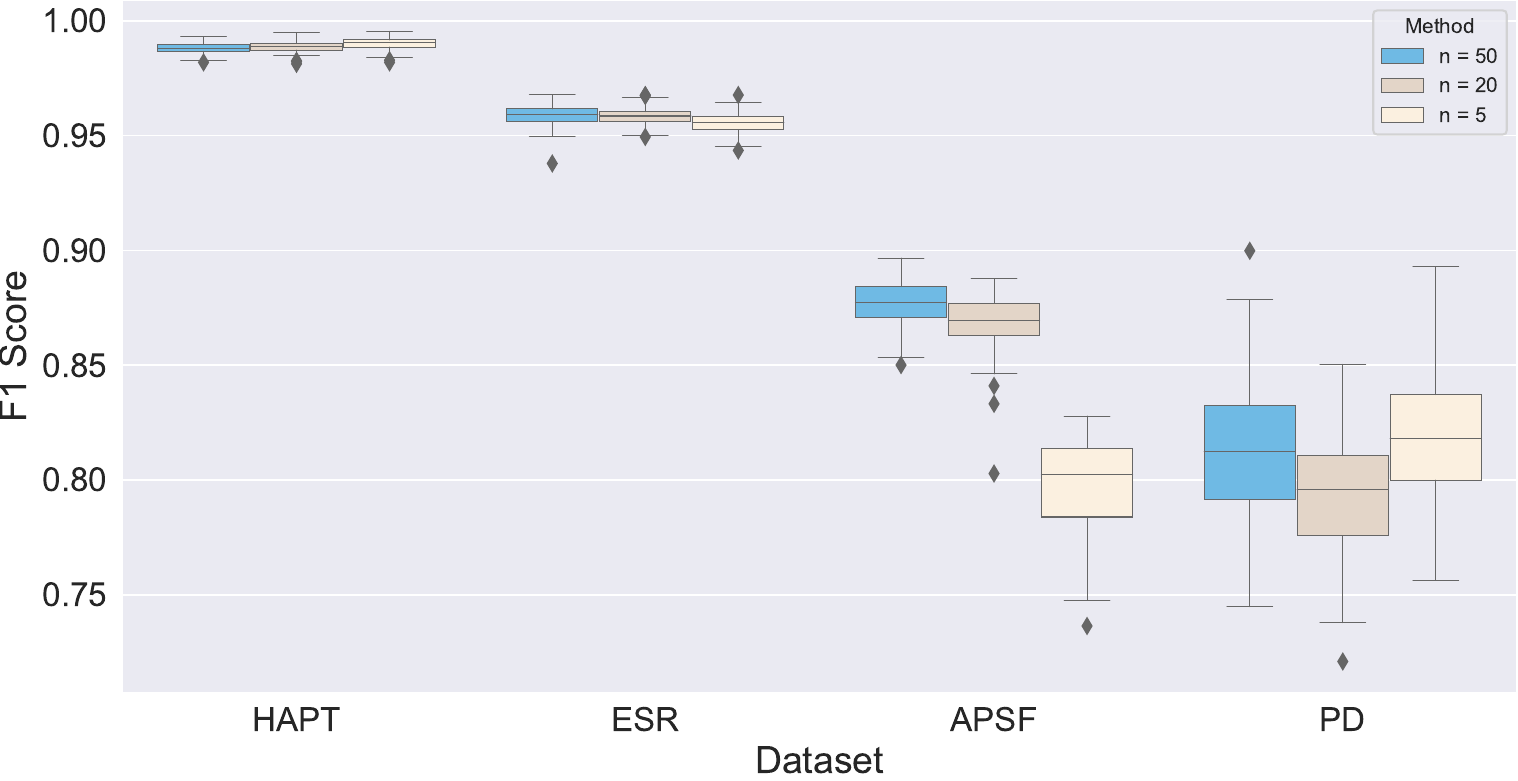}
  \caption{Ablation study results for n 5, 20, 50.}
  \label{fig:ablation_boxplot}
\end{figure}

As mentioned previously, the $n$ (multiplier of $\sigma$) has been fixed at a value of 50 in this investigation. This particular selection is motivated by the fact that the normalized data often yields a diminutive standard deviation. By assigning a larger figure for standard deviation, it ensures sufficient perturbation of the features. Nonetheless, it raises a pertinent question about the influence of this coefficient on the efficacy of the proposed methodology. Table \ref{tab:Ablation} and Figure \ref{fig:ablation_boxplot} displays the results of the proposed method with 'n' values set to 5, 20, and 50. It should be noted that during this analysis the other parameters, including the shallow learner structure, the number of iterations, and epochs, were kept constant throughout the analysis. The result obtained from 100 evaluations after feature selection (similar to previous section).\\
From the results of the ablation study, it can be inferred that increasing 'n', and therefore the perturbation, influences the feature selection process and subsequently the ANN's performance in different ways across the datasets.

For instance, the 'Smartphone-Based Recognition of Human Activities' dataset demonstrates that a greater perturbation might lead to more stringent feature selection, resulting in fewer features being selected while maintaining similar performance levels. This could suggest that larger perturbations help to highlight only the most influential features, as minor ones might be 'washed out' due to the higher noise levels.

In the 'Failure at Scania Trucks' and 'Epileptic Seizure Recognition' datasets, an increase in 'n' appears to reveal more features that contribute to the performance of the ANN, indicating that a greater degree of perturbation might be useful in uncovering hidden or complex relationships in the data.

However, the results from the 'Parkinson's Disease (PD) classification' dataset provide a nuanced view, suggesting that there might not be a linear relationship between the magnitude of perturbation and the performance of the ANN. Here, the number of selected features and the F1 score do not demonstrate a consistent trend with increasing perturbation, highlighting the intricacies of the ANN's response to perturbations in this context.

Thus, while the perturbation multiplier 'n' clearly impacts the ANN's behavior, the nature and extent of this impact can vary greatly based on the dataset's inherent properties. This underscores the importance of fine-tuning 'n' based on specific dataset characteristics to optimize the ANN's performance.

Overall, the proposed method has proven to be capable of selecting crucial features even with variations in $n$. The comparison with the number of features selected by the Boruta algorithm also demonstrates that it continues to select fewer features. In other words, the proposed method consistently outperforms the Boruta algorithm with respect to the quantity of selected features.  This aligns with the objective of feature selection, which is to select the minimum possible number of features while still maintaining adequate performance in modeling the response variable.

\section{Conclusion}\label{section:Conclusion}
The innovation of this method lies in the intentional modification of shadow features' characteristics, differing from traditional approaches where shadow features retain the same statistical properties as their original counterparts. Therefore, this study proposed a new variant of the Boruta, called Nnoise-augmented Boruta. In light of the comprehensive evaluation conducted in this study, it can be conclusively stated that the proposed noise-augmented Boruta methodology offers substantial improvements over the classic Boruta algorithm, where the proposed model consistently outperforms in terms of selecting fewer but more essential features across multiple datasets. This performance adheres to the fundamental principle of feature selection: reducing model complexity while preserving predictive power. \\
Moreover, the conducted ablation study provides valuable insights into the role and impact of the standard deviation multiplier 'n' within the proposed methodology. The multiplier, by influencing the perturbation, demonstrates substantial control over the feature selection process and subsequent performance of the Artificial Neural Network. Importantly, this relationship is not linear, and the specific characteristics of the dataset strongly influence the optimal value for 'n'. In conclusion, the proposed noise-augmented Boruta methodology presents a promising advance in the domain of feature selection. Its superior performance, coupled with the insightful findings from the ablation study, demonstrates its potential for broad applicability across various machine-learning tasks. However, careful tuning of its perturbation parameter 'n' is critical to ensure optimal results, emphasizing the need for a context-specific approach when applying this technique. 
The possible direction to extend this work is incorporating uncertainty metrics. This would pivot the focus towards not just discerning features that decrease model performance when perturbed, but also understanding the model's certainty regarding such perturbations.

\ifCLASSOPTIONcaptionsoff
  \newpage
\fi

\bibliographystyle{IEEEtran}
\bibliography{Refs}


\vfill

\end{document}